\documentclass{article}
\usepackage{spconf,amsmath,graphicx}
\usepackage{times}
\usepackage{epsfig}
\usepackage{graphicx}
\usepackage{amsmath}
\usepackage{amssymb}
\usepackage{booktabs}
\usepackage{subcaption}
\usepackage{kantlipsum} %<- For dummy text
\usepackage{mwe} %<- For dummy images
\usepackage{floatrow}
\usepackage{multirow}
\usepackage[table,x11names]{xcolor}
\usepackage[utf8]{inputenc}
\usepackage{siunitx}
\usepackage{caption}
\usepackage{multicol}
\usepackage{xcolor, soul}
\sethlcolor{lightgray}
\usepackage{xspace}

\def\modelone{{AT-GDINO-SAM}}
\def\modelthree{{SAM-BIND}}
\def\modeltwo{{OWOD-BIND}}
\def\tf{{\em an untrained}}
\newcommand{\eg}{e.g.\xspace}
\newcommand{\ie}{i.e.\xspace}

\usepackage[breaklinks=true,bookmarks=false]{hyperref}
\title{Leveraging Foundation Models for Unsupervised Audio-Visual Segmentation}

\name{
Swapnil Bhosale, Haosen Yang, Diptesh Kanojia, Xiatian Zhu
}
\address{University of Surrey, UK}

\begin{document}
%\ninept
%
\maketitle
\begin{abstract}
Audio-Visual Segmentation (AVS) aims to precisely outline audible objects in a visual scene at the pixel level.
%setting at the pixel level, specifically by identifying the segmentation mask of these auditory objects. 
Existing AVS methods require fine-grained annotations of audio-mask pairs in supervised learning fashion.
%for guidance. 
This limits their scalability 
since it is time consuming and tedious to acquire such 
cross-modality pixel level labels.
To overcome this obstacle,
in this work we introduce {\bf\em unsupervised audio-visual segmentation}
with no need for task-specific data annotations and model training.
For tackling this newly proposed problem, we formulate a novel {\bf\em Cross-Modality Semantic Filtering} (CMSF) approach to accurately associate the underlying audio-mask pairs by leveraging the off-the-shelf multi-modal foundation models (\eg, detection 
\cite{gong2021ast}, open-world segmentation \cite{kirillov2023segment} and multi-modal alignment \cite{girdhar2023imagebind}).
Guiding the proposal generation by either audio or visual cues, we design two training-free variants: \modelone{} and \modeltwo{}.
Extensive experiments on the AVS-Bench dataset show that our unsupervised approach can perform well in comparison to prior art supervised counterparts across complex scenarios with multiple auditory objects.
Particularly, in situations where existing supervised AVS methods struggle with overlapping foreground objects, our models still excel in accurately segmenting overlapped auditory objects. 
Our code will be publicly released.
\end{abstract}

\section{Introduction}
\label{sec:intro}

In the realm of audio-visual segmentation (AVS), the challenge lies in delineating sounding objects that align with audio cues present within video sequences. Historically, the intersection of audio-visual signals has been studied through the lens of self-supervised learning~\cite{afouras2020self, rouditchenko2019self}. However, this methodology exhibits constraints, especially in real-world applications that necessitate precise segmentations, such as in video surveillance, integrated video editing, and advanced robotics. A recent study~\cite{zhou2022audio} approached the AVS quandary using supervised learning and developed a manually annotated video dataset, emphasizing pixel-level segmentations of audio-responsive objects. Yet, extensions of this method \cite{chen2023closer, liu2023annotation, hao2023improving, liu2023bavs, zhou2022contrastive, mao2023contrastive, zhou2023audio}, are not without its shortcomings: (1) substantial manual audio-mask annotation overheads, (2) a dataset spanning a narrow range of categories, and (3) intensive training requisites.

As the landscape of foundational frameworks within a specific domain expands, as exemplified by works like ~\cite{kirillov2023segment, liu2023grounding, girdhar2023imagebind}, a compelling question arises: Can their accumulated insights be coalesced for the enhancement of the AVS task? To answering this, we present \textit{Cross-Modality Semantic Filtering (CMSF)}, an advanced cascade of foundational models that employs \tf{} approach. 
This approach harmoniously integrates knowledge from diverse pre-training paradigms, setting a new benchmark for a resilient and adaptive AVS methodology.
Particularly, we propose two mechanisms under CMSF bootstrapping foundation models pertaining to uni-modalities and related tasks such as segmentation, audio tagging and grounding. In each of the mechanism we alternate between the audio and visual modality, wherein one is exploited to generate proposal masks, whereas the other is used as guidance to condition/filter among segmentation masks/bounding box proposals. As shown in Fig. \ref{fig:pipe-1}, our first mechanism, \modelone{} utilizes audio cues to generate audio tags and eventually proposal segmentation masks. Along the same lines, we also propose \modeltwo{} which utilizes the visual cue to generate proposal bounding boxes which are further filtered based on their cosine similarity with the audio counterpart, both projected through into shared latent space learned by the ImageBIND model \cite{girdhar2023imagebind}. 

\subsection{Supervised AVS approaches}
Current methods for addressing the AVS task involve training a fully supervised model responsible for identifying all pixels associated with audible visual elements within a scene. The current AVS-Benchmark, TPAVI \cite{zhou2022audio} takes a different approach by employing multiple attention fusion modules, establishing a direct link between the encoder and decoder to learn the alignment between audio and visual components. Another approach AV-SAM \cite{mo2023av} makes use of pixel-level audio-visual fusion from SAM's pre-trained image encoder. The amalgamated cross-modal features are then directed into SAM's prompt encoder and mask decoder, generating audio-visual segmentation masks.

We contend that the supervised training paradigm for AVS necessitates resource-intensive annotations 
%for producing pixel-level binary masks. 
and the effectiveness of existing AVS systems, especially when confronted with multiple audible objects, faces limitations due to scene diversity, along with the quantity and quality of annotated audio masks. To address these challenges, we introduce an alternative approach that obviates the requirement for audio-mask pairs by capitalizing on \tf{} methodology that incorporates insights from established foundational models.

\section{Methodology}
Given a video sequence $\{I_t\}_{t=1}^{T} \in \mathcal{R}^{H \times W}$ with $T$ non-overlapping continuous image frames where $H$ and $W$ denote the height and width of each frame, along with an audio sequence $A$ with the same temporal resolution as $I$, $\{A_t\}_{t=1}^{T}$, the goal of an AVS system is to generate segmentation masks $\{G_t\} \in \mathcal{R}^{H \times W}$. The masks are interpreted as pixel-wise labels which highlight the sounding source object in $I_t$ that is closest probable source of the sound in $A_t$. In contrast to the existing AVS approaches, which utilize a supervised training pipeline incorporating pre-annotated audio-mask pairs, our work is motivated towards \tf{} pipeline for AVS which exploits the publicly available models pre-trained on foundational tasks namely, audio tagging, object detection and segmentation. These models have been trained on the open-set setting, with the latter two supporting zero-shot inference, and are hence aligned to our use-case.
\vspace{-1em}
\subsection{Pre-training Model}
\vspace{-0.7em}
%\paragraph{Audio Tagging (AT)}
%\vspace{-1em}
\textbf{Audio Tagging (AT): }An Audio Tagging (AT) model assigns descriptive tags to audio clips, representing features like musical instruments, environmental sounds, and more. Current research in AT has shifted from hand-crafted to end-to-end models, evolving from CNNs \cite{kong2020panns} to convolution-free attention models like the Audio Spectrogram Transformer (AST) \cite{gong2021ast}. We adopt AST to generate audio tags for the AVS task, utilizing its capacity to capture global context in audio sequences with multiple events. Trained on Audioset \cite{liu2023grounding}, AST detects diverse polyphonic audio events across 521 categories like animal sounds, instruments, and human activity.

\noindent\textbf{Open World Object Detector (OWOD): }
%\vspace{-1em}
%Class-agnostic 
%text-prompt : Grounding DINO
Traditional object detection treats unfamiliar objects as background, while open-world detection allows models to handle unknown objects during training and inference \cite{bendale2015towards, dhamija2020overlooked}.
%The aim is to recognize and learn about these unknowns incrementally.
\cite{joseph2021towards} introduced the open-world detector (ORE) with 80 COCO \cite{lin2014microsoft} classes split into 4 tasks, adding 20 classes to known categories in each.
%They used a Faster-RCNN \cite{ren2015faster} base detector and a class-agnostic region proposal network (RPN) for unknown objects. 
\cite{maaz2022class} extended this using class-agnostic proposals from a Multiscale Attention ViT \cite{dosovitskiy2020image} with late fusion (MAVL).
%, showing that multimodal ViTs trained on image-text pairs can adapt well to new domains and novel objects. 
In our work, we use this model to generate class-agnostic object proposals that are further processed by the segmentation network. 

\noindent\textbf{Segment-Anything Model (SAM): }
%\vspace{-1em}
The Segment-Anything model \cite{kirillov2023segment}, trained on a vast dataset of 1 billion masks, is renowned for its prompt-based zero-shot segmentation capabilities as well as its versatility as a foundation model to merge cross-modal representations using multimodal prompts \cite{zhang2023survey}.
%SAM excels at zero-shot object segmentation. 
%Recent research highlights SAM's versatility as a foundational model, using learnable multimodal prompts to merge cross-modal representations \cite{zhang2023survey}. 
SAM's core includes a prompt encoder capable of handling varied prompt types like points, bounding boxes, and text. We extend SAM's segmentation capabilities by leveraging visual prompts from OWOD to generate pixel-level masks.

\subsection{Cross-Modality Semantic Filtering (CMSF): }
\vspace{-0.7em}
%In this section, we explain our proposed pipelines for training-free AVS. 
The core of the AVS task is associating audio content with object spatial arrangement. This involves processing cross-modal input cues: acoustic and visual, wherein one is used to generate proposals while other one is used to filter/condition the available proposals.% We introduce two methods based on the utilized cue below. 
%First, \modelone{} leverages audio cues to craft proposal masks, capturing frame-specific spatial details. Second, \modeltwo{} employs the visual cue for creating proposal bounding boxes, guided by the audio cue.
% The fundamental essence of the AVS task entails linking audio content details with the spatial arrangement of objects. This indicates the presence of two input indicators that necessitate processing: the acoustic cue and the visual cue. We introduce two methodologies that depend on which cue is employed to facilitate this mapping in each modality. To elaborate, we initially present \modelone{}, which capitalizes on audio cues to create proposal segmentation masks, harnessing the spatial particulars within the frame. Secondly, we introduce \modeltwo{}, which employs the visual cue to construct proposal bounding boxes while utilizing the audio cue as its guiding reference.
%processes the visual cue to extract conditioning information which is used to match the audio content.  
%\vspace{-1em}

\begin{figure*}[t]
    \centering
    \includegraphics[width=0.69\textwidth]{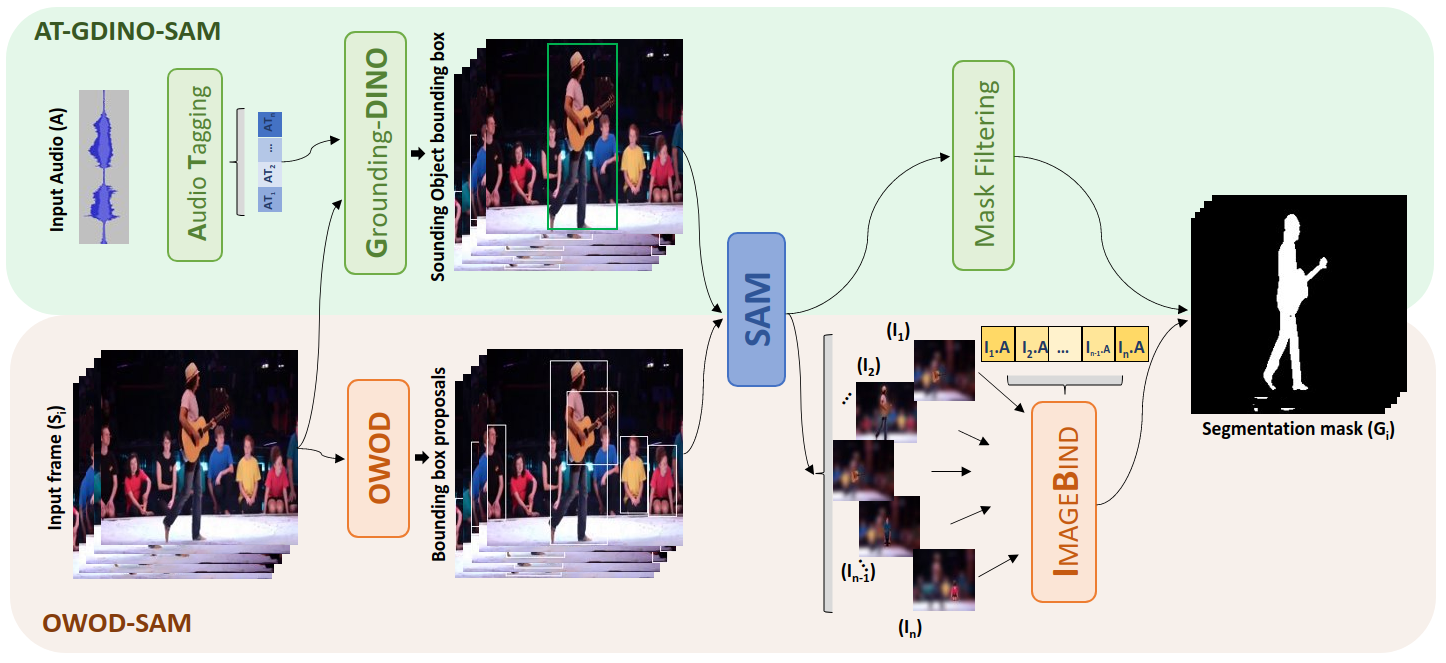}
    \caption{\textbf{Overview of CSMF}: Bounding box proposals are generated using a cascade of two pre-trained models: Audio Tagging and GroundingDINO. Favourable boxes are passed as input to SAM to yield segmentation masks.}
    \label{fig:pipe-1}
\end{figure*}

\noindent\textbf{Generating proposals using audio cue: \modelone{}}
%\vspace{-1em}
%First, 
Each audio sequence, $A_t$ is padded to a maximum of 960 msec and processed by a pre-trained AST model, yielding audio tags, $\{AT_{i}\}_{i=1}^{C_a}$, which are ranked based on their probability scores across 521 generic classes from the Audioset. Relevant audio tags are filtered using an empirically determined threshold, $\tau_{AT}$ and forwarded to a pre-trained GroundingDINO model \cite{liu2023grounding}, generating bounding boxes in image frame $I_t$. These boxes serve as visual prompts for SAM \cite{kirillov2023segment}, producing binary masks (see Fig. \ref{fig:pipe-1}).
%. The aggregating of masks associated with filtered audio tags forms the final segmentation mask 

\noindent\textbf{Generating proposals using visual cue: \modeltwo{}}
%To create proposal boxes using the visual cue, we use the pre-trained OWOD model. 
Given an image frame, $I_t$, we use OWOD to generate proposal bounding box proposals, $\{BB_{i}\}_{i=1}^{C_v}$ with $C_v$ proposals. These are filtered by objectness score \cite{maaz2022class} using threshold $\tau_{BB}$. To link boxes and acoustic cues, both modalities need shared latent space embedding semantics. Towards this end, we utilize ImageBIND's \cite{girdhar2023imagebind} latents, and extract image and audio embeddings which are used to rank the proposals by cosine similarity with audio embedding. Bounding boxes above $\tau_{BIND}$ form the final mask.
%The detailed architecture is provided in Fig. \ref{fig:pipe-2}. 

As an alternative approach, instead of relying on OWOD to generate bounding box proposals, we can randomly position single-point input prompts in a grid across the image. From each point, SAM can predict multiple masks. These masks are refined and filtered for quality, employing non-maximal suppression (NMS) to remove duplicates. We term this approach \modelthree{}.

\section{Experiments and Results}
\vspace{-0.7em}
\noindent\textbf{Dataset: }
To assess our proposed methods, we use the recently released AVSBench dataset \cite{zhou2022audio}. AVSBench contains videos from YouTube, split into 5 distinct clips. Each clip is paired with a spatial segmentation mask indicating the audible object. AVSBench includes two subsets: the Semi-supervised Single Source Segmentation (S4) and the fully supervised Multiple Sound Source Segmentation (MS3), differing in the number of audible objects.

\noindent\textbf{Metrics: }
Motivated towards an Unsupervised AVS approach, we compare our proposed pipelines on S4 and MS3 test splits without using any audio-mask pairs from train/validation. Similar to existing supervised methods, we use average Intersection Over Union ($M_{IOU}$) and $F_{score}$ as metrics. Higher $M_{IOU}$ implies better region similarity, and elevated $F_{score}$ indicates improved contour accuracy.

\noindent\textbf{Implementation Details: }
We resize of all image frames to size 224 $\times$ 224. 
%Unless stated otherwise, 
For all our evaluation results we use $\tau_{AT}$ as 0.5 for \modelone{} and $\tau_{BB}$ and $\tau_{BIND}$ as 0.5 and 0.7 respectively, for \modeltwo{}. In \modelthree{}, we use IOU threshold of 0.5 for NMS.
\subsection{Quantitative Comparison}
\vspace{-0.7em}
We present our primary AVSBench results on the test set for both S4 and MS3 in Table \ref{tab:main_results}. We can observe that both, \modeltwo{} and \modelthree{} outperform \modelone{} in terms of both $M_{IOU}$ and $F_{score}$ by a significant margin.
It should be noted that AST achieves a mean average precision (mAP) of only 0.485 on the Audioset dataset, and hence the generated audio tags are highly prone to errors. Additionally, we believe despite AST's training data \ie, Audioset follows a generic ontology, many rare events such as "lawn mover", "tabla" etc. are under-represented and hence are unable to cope with an open-set inference. %Furthermore, we believe there is an inherent mismatch between the audio tags  predicted by a pre-trained AT model which are often not completely indicative of the object to be segmented, for instance .
Between \modeltwo{} and \modelthree{}, we can observe that \modeltwo{} achieves an absolute improvement of over 0.06 and 0.08 in terms of $M_{IOU}$ and $F_{score}$ respectively under the MS3 setting. Similar trend is observed when comparing both the approaches under the S4 setting, with 0.16 improvement in both $M_{IOU}$ and $F_{score}$ respectively.

\begin{table}[h]
    \centering
    \caption{Performance comparison of our proposed \tf{} pipelines on the AVSBench \texttt{test} split. \hl{Grayed}: Using explicit audio-visual mechanism \ie, supervised learning.}
    \begin{tabular}{c|cc|cc}
    
    \toprule
    Approach & \multicolumn{2}{c}{S4}&\multicolumn{2}{c}{MS3}\\
     &$\rm{M_{IOU}}$&$\rm{F_{score}}$&$\rm{M_{IOU}}$&$\rm{F_{score}}$ \\
    \midrule\midrule
    \rowcolor{lightgray} AVSBench \cite{zhou2022audio} & $0.78$ & $0.87$ & $0.54$ & $0.64$ \\
    %\midrule
    \rowcolor{lightgray} AV-SAM \cite{mo2023av} & $0.40$ & $0.56$ & - & - \\
    \midrule
    \modelone{} & $0.38$ & $0.46$ & $0.25$ & $0.29$  \\
    %\midrule
    \modelthree{} & $0.42$ & $0.51$ & $0.28$ & $0.36$ \\
    %\midrule
     \modeltwo{} & \bf{0.58} & \bf{0.67} & \bf{0.34} & \bf{0.44} \\
    \bottomrule
    \end{tabular}
    \label{tab:main_results}
\end{table}

\begin{figure}[H]
    \centering
    \includegraphics[width=0.89\textwidth]{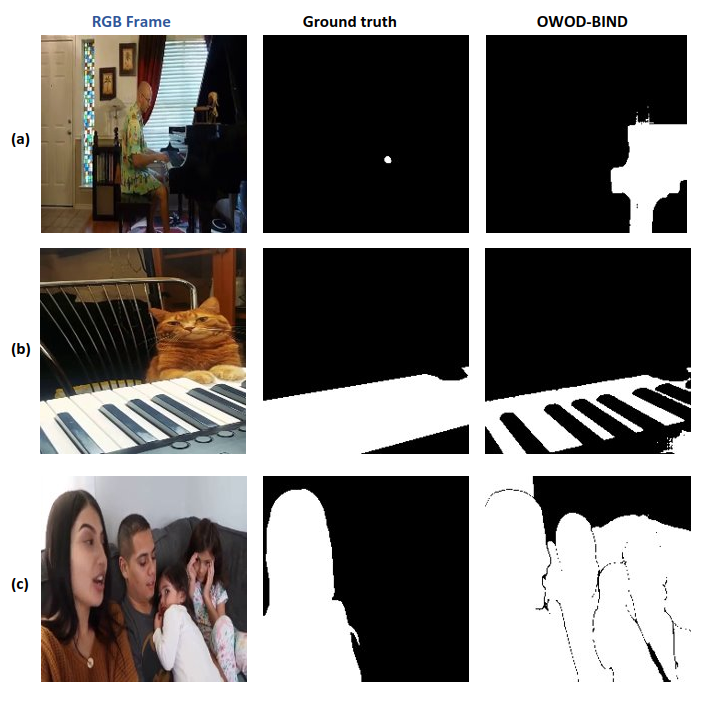}
    \caption{Unsuccessful outcomes: (a) Imprecise localization (b) Over-segmentation (c) Improper activity detection.}
    \label{fig:fail_cases}
\end{figure}

\noindent\textbf{Comparison with supervised AVS approaches: }
%In table \ref{tab:comp_sup_avs}, we compare our best performing framework, \modeltwo{} with existing supervised AVS approaches. 
We propose \tf{} paradigm without any use of audio-mask pairs for fine-tuning existing foundation models. We fall short of the AVS-Benchmark by 0.20 in terms of both $M_{IOU}$ and $F_{score}$, when evaluating under the MS3 setting, where the latter benefits from the effective audio-visual contrastive loss design and trainable parameters. Upon careful invigilation of the generated masks, we observe that beyond the lack of trainable parameters, the proposed approach performs poorly primarily because of the following reasons, (a) inability of SAM to perform precise localization to the level depicted in the annotated ground-truth masks. In Fig. \ref{fig:fail_cases}(a) \modeltwo{} is able to delineate piano from other objects $\{$human, portrait etc$\}$, however AVSBench requires solely the keys of the piano to be highlighted.
(b) the SAM performs over-segmentation particularly, when the objects of interest are zoomed in. Fig. \ref{fig:fail_cases}(b) shows that SAM tends to segment the white keys from the black keys, while AVSBench expects the entire piano including all the keys to be highlighted under one single object mask. 
(c) \modeltwo{} performs segmentation of isolated frames without capturing the temporal aspect within consecutive frames and hence lacks the propagated activity detection. In Fig. \ref{fig:fail_cases}(c) the input audio, $A_t$ consists of human speech, however \modeltwo{} fails to distinguish between the speaking and non-speaking humans, since ImageBIND associates each detected person with human speech presented in $A_t$.
\vspace{-1em}
\subsection{Qualitative comparison}
\begin{figure}[H]
    \centering
    \subfloat[]{\includegraphics[width=0.9\textwidth]{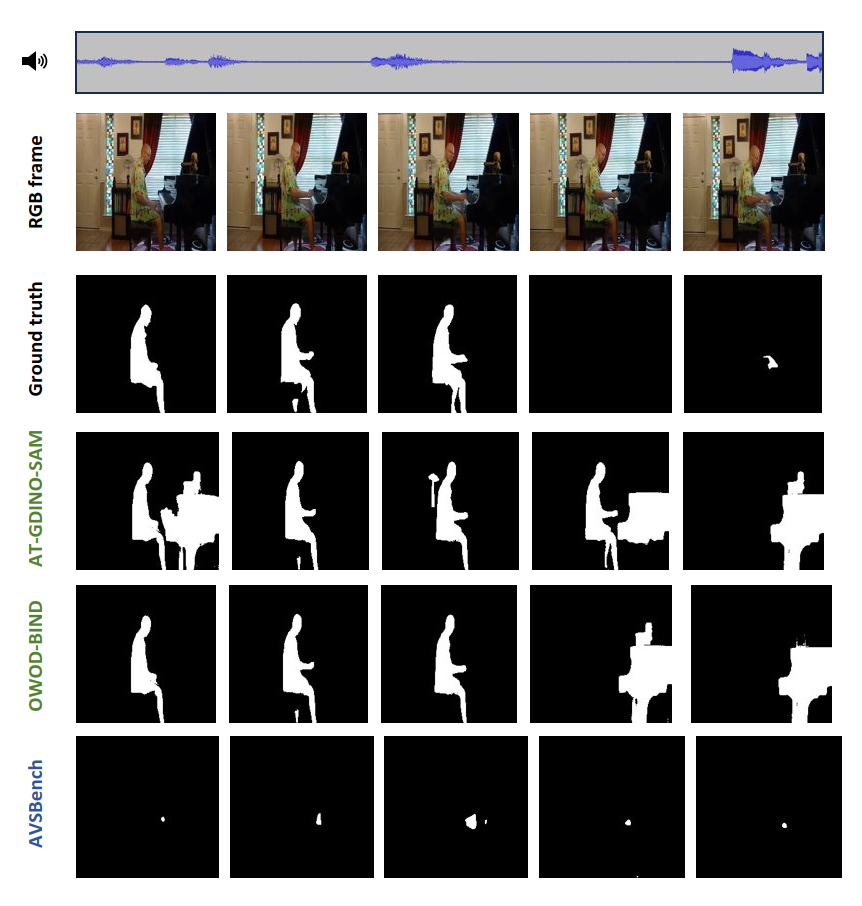}}
    
    \subfloat[]{\includegraphics[width=0.9\textwidth]{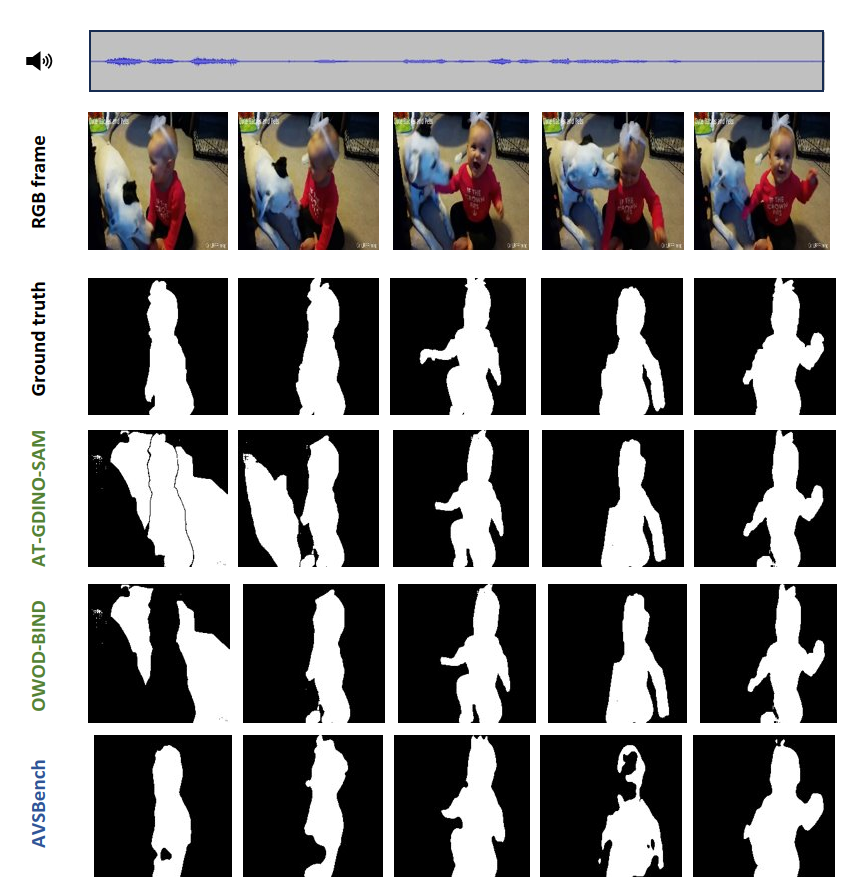}}
    
    \caption{Qualitative comparisons with AVSBench and ground-truth segmentation masks. \modeltwo{} produces more precise segmentation of overlapping objects without explicit audio-visual fusion.}
    \label{fig:avs_preds}
\end{figure}

\vspace{-1em}
We demonstrate the qualitative comparison of our proposed approach with the AVS-Benchmark in terms of the quality of the generated masks. We observe that although, \modeltwo{} results in a lower average $M_{IOU}$ and $F_{score}$ against the AVS-Benchmark, it still generates more finer details especially when segmenting non-cubical objects for instance, humans. In Fig. \ref{fig:avs_preds}, we highlight that despite lacking an explicit audio-visual fusion mechanism, our proposed approach is able to alternate among multiple sounding objects in the foreground $\{$\textit{human, piano}$\}$. It can be observed that the OWOD foundation model helps generate continuous masks despite having multiple granular background objects. AVS-Benchmark partially localizes the sounding object however, generates discontinuous masks, troubled by the presence of spatially overlapping foreground objects $\{$\textit{dog, baby}$\}$.

% \begin{figure*}[ht]
%     \centering
%     \includegraphics[width=0.99\textwidth]{avs_pred.png}
%     \caption{Qualitative comparisons with AVSBench and ground-truth segmentation masks. \modeltwo{} produces more precise segmentation of overlapping objects without explicit audio-visual fusion.}
%     \label{fig:avs_preds}
% \end{figure*}

\vspace{-1em}
\section{Conclusion}
\vspace{-1em}
In this study, we approach Audio-Visual Segmentation (AVS) unconventionally, using unsupervised learning and insights from foundational multimodal models. We present a novel Cross-modal Semantic Filtering paradigm that calls for a training-free approach to AVS. This marks the first AVS exploration without pre-annotated audio-mask pairs. Our approach achieves accurate segmentation of overlapping masks without explicit audio-visual fusion. We validate the efficacy of open world foundation models in precisely distinguishing auditory elements within visual contexts, when compared to existing supervised methods. Future work involves integrating temporal context and mitigating SAM model over-segmentation via stricter audio guidance, all while maintaining an unsupervised framework.

\vfill\pagebreak

\label{sec:refs}
\bibliographystyle{IEEEbib}
\bibliography{strings,refs}

\end{document}